\documentclass[10pt,conference]{IEEEtran}

\usepackage{amsmath,amsfonts,bm}

\def\eqref#1{equation~\ref{#1}}

\def\1{\bm{1}}

\DeclareMathAlphabet{\mathsfit}{\encodingdefault}{\sfdefault}{m}{sl}
\SetMathAlphabet{\mathsfit}{bold}{\encodingdefault}{\sfdefault}{bx}{n}

\usepackage{subcaption}
\usepackage{hyperref}
\usepackage{multirow}
\usepackage{url}
\usepackage[utf8]{inputenc} 
\usepackage[T1]{fontenc}    
\usepackage{mdframed}
\usepackage{hyperref}       
\usepackage{url}            
\usepackage{booktabs}       
\usepackage{amsfonts}       
\usepackage{amsmath}
\usepackage{nicefrac}       
\usepackage{microtype}      
\usepackage{xcolor}         
\usepackage{graphicx}
\usepackage{multicol}
\usepackage{algorithm}
\usepackage{algpseudocode}

\author{
\IEEEauthorblockN{Tiago F. Tavares}
\IEEEauthorblockA{
Department of Computer Science\\
Insper\\
São Paulo, Brazil\\
tiagoft1@insper.edu.br
}
\and
\IEEEauthorblockN{Fabio Ayres}
\IEEEauthorblockA{
Department of Computer Science\\
Insper\\
São Paulo, Brazil\\
fabioja@insper.edu.br
}
\and
\IEEEauthorblockN{Paris Smaragdis}
\IEEEauthorblockA{
Massachusetts Institute of Technology\\
Boston, MA, USA\\
paris@mit.edu
}
}

\begin{document}

%
\title{Diagnosing Neural Convergence with Topological Alignment Spectra}
%
%
%



\markboth{Journal of \LaTeX\ Class Files,~Vol.~14, No.~8, August~2015}%
{Shell \MakeLowercase{\textit{et al.}}: Bare Demo of IEEEtran.cls for IEEE Journals}


\maketitle
\begin{abstract}


Representational similarity in neural networks is inherently scale-dependent, yet widely used metrics such as Centered Kernel Alignment (CKA) and Procrustes analysis provide only global scalar estimates. These scalars often fail to distinguish micro-scale geometric jitter (local noise) from macro-scale semantic reorganization, compressing multi-scale structural relationships into a single uninformative value. We introduce the Topological Alignment Spectrum (TAS), a multi-scale diagnostic tool that sweeps normalized mean Jaccard similarity over varying neighborhood sizes. By normalizing the metric over an analytically-derived expected range (from expected overlap under randomness to perfect alignment), TAS yields a dimension-invariant metric over a spectrum of scales, where one indicates perfect structural alignment, zero reflects chance-level agreement, and negative values signal active anti-alignment at specific scales. Experiments on synthetic point clouds demonstrate that TAS allows the recognition of  distinct types of alignment perturbation: local jitter harms fine-grained neighborhoods but preserves cluster-level structure, while cluster-center shuffling preserves local similarity but disrupts global alignment -- phenomena that remain invisible or conflated under global, single-scalar metrics. Applying TAS to the MultiBERTs collection reveals that fine-tuning induces comprehensive topological reorganization across scales, challenging the view of task adaptation as merely conservative or localized. While models from different random seeds remain locally divergent, semantic clusters emerge as the dominant scale of alignment. TAS thus offers a granular, topology-aware alternative for diagnosing convergence and representational stability in deep networks.
        
\end{abstract}


\section{Introduction}
\label{sec:introduction}
Measuring the similarity between neural networks is becoming increasingly relevant to the science of deep learning. Beyond mere performance benchmarking, representational similarity analysis allows researchers to assess whether distinct models converge to universal conceptual structures, predict transferability, and audit model reliability in the absence of ground truth labels~\cite{klabunde2025}. While functional metrics (based on output behavior) are often the end goal, recent findings suggest that functionally similar models can exhibit vastly different internal topologies, necessitating robust representational metrics to diagnose the true structural alignment of latent spaces.

This disconnect between internal geometry and external behavior is critical because, while accuracy remains the gold standard for benchmarking, recent literature reveals that functional equivalence often masks deep representational instability~\cite{black2022}. Modern neural networks have been demonstrated to be \textit{underspecified}~\cite{damour2020underspecificationpresentschallengescredibility}, meaning they can converge to topologically distinct configurations that yield identical test accuracy but exhibit conflicting predictions on out-of-distribution data~\cite{hamman2025quantifyingpredictionconsistencyfinetuning}. This has motivated work investigating geometries of changes within neural networks in situations such as fine-tuning~\cite{zhou-srikumar-2022-closer,merchant-etal-2020-happens}.

However, representational divergence is rarely uniform. Distortions between embedding spaces can manifest in distinct forms -- ranging from global affine transformations and rotations to localized stochastic noise (``jitter'') -- often occurring simultaneously but dominating at different geometric scales. The vast majority of current methods for similarity evaluation (such as CKA or PWCCA) compress this high-dimensional relationship into a single global scalar~\cite{klabunde2025}. We argue that such compression inherently loses information regarding the \textit{granularity} of the alignment.

To fully characterize structural consistency, we propose that one must analyze similarity not as a point estimate, but as a continuous spectrum across varying neighborhood sizes. This multi-scale perspective is essential to disentangle distinct topological phenomena, allowing us to determine whether models diverge due to micro-scale geometric instability or macro-scale semantic reorganization.

The source code for the experiments presented here is available at \url{https://github.com/InsperML/pointcloudsimilarity}.

\section{Related work}
\label{sec:related}

Items from a dataset can have different representations. The similarity between representations can be measured with a \emph{similarity index} $s(X,Y)$, which compares different representations $X \in \mathbb{R}^{n \times p_1}$ and $Y \in \mathbb{R}^{n \times p_2}$ (with possibly different dimensionalities $p_1$ and $p_2$, with $p_1 \leq p_2$ without loss of generality) of the same set of $n$ items~\cite{hintoncka2019}. 

Arguably, a similarity index should be invariant to orthonormal linear transformations and isotropic non-zero scaling and translation~\cite{hintoncka2019}. More generally, a similarity index could be invariant to general invertible linear transformations~\cite{raghu2017}, but this property will not be adopted in the present work as it implies that all datasets have the same similarity index if $p_1>n$~\cite{hintoncka2019}.

\subsection{Procrustes}
The Procrustes problem consists on finding a linear transform that maps $X$ onto $Y$~\cite{Gower1975} and calculating the error in this mapping. Procrustes analysis is fundamentally limited because it required both point clouds to have equal dimension. Despite its limitations, Procrustes (in particular, its orthogonal variant~\cite{schonemann1966generalized}) has been shown to correctly identify representation alignments in a diversity of modern language models~\cite{maystre2025embeddingmodelsmeetprocrustes}.

\subsection{Canonical Correlation Analysis}
Canonical Correlation Analysis (CCA) aims to find two linear transforms $Q_x$ and $Q_y$ so that the correlation between $XQ_x$ and $YQ_y$ is maximized~\cite{Hotelling1992}, which allows analyzing point clouds in different dimensions. CCA has been observed to be sensible to noise, which lead to the incorporation of Singular Value Decomposition in SVDCCA~\cite{raghu2017} to mitigate noise in low-variance directions. Subsequently, the introduction of a projection-weighting scheme in Projection-Weighted CCA (PWCCA) allowed to account for the differences in the contributions of each direction towards the representation's information content~\cite{morcos2018}.

\subsection{GULP: GULP is Uniform Linear Probing}
A different approach, called GULP~\cite{GulpNEURIPS2022}, defines similar representations as those that lead to the same predictive performance for a set of items. 

Following a similar idea, ContraSIM~\cite{Rahamim2023} relies on the idea that it should be possible to map similar representations to a common embedding space using contrastive loss. Thus, ContraSIM~\cite{Rahamim2023} uses the loss of such mapping as a similarity measure.

We observe that both GULP and ContraSIM can be seen as extensions of the Canonical Correlation Analysis (CCA)~\cite{Hotelling1992}, as they measure the error obtained due to projecting the representations $X$ and $Y$ onto a common space. ContraSIM~\cite{Rahamim2023} allows greater flexibility in this projection by allowing the use of arbitrary encoders, but this also implies in the need to train an encoder. Training contrastive losses implies in a series of particular problems and, because of that, ContraSIM was not used in further experiments.

\subsection{Optimal Transport}
Optimal Transport theory brings forward a different idea for point cloud similarities based on the Gromov-Wasserstein (GW) distance~\cite{Mmoli2011}. It seeks for a probabilistic transport plan (that is, a probabilistic mapping matrix) that minimizes the discrepancies between the point-wise distances within $X$ and $Y$. In contrast to CCA-based methods, it measures how much \textit{distances} are preserved, even if points are shuffled from $X$ to $Y$.

\subsection{Centered Kernel Alignment}
An important measure for similarity between representations is the Centered Kernel Alignment (CKA) measure~\cite{hintoncka2019}.
Similarly to Optimal Transport, CKA draws from the idea of assessing how distances between items change from one representation to another. With a linear kernel, changes in the distance between point clusters have a larger impact than disparities on the distance between points within each cluster, because changes in the separation of whole clusters are usually larger in magnitude than local changes. By using an RBF kernel ($\exp{-D(X)/2\sigma^2}$, where $D(X)$ is a matrix with pairwise-distances between the items of $X$) it is possible to give more weight to changes made on smaller distances.

However, the weighting process is distance-based, that is, the choice of an adequate value for $\sigma$ depends on the radius of the localities being analyzed. This value is usually hard to consistently obtain, as it can change even if data is subject to isotropic transformations. For this reason, $\sigma$ can be obtained as a fraction $\alpha$ of the median distance between points within the cloud~\cite{hintoncka2019}.

\subsection{Representation Topology Divergence}

Another possible definition for similarity lies on the analysis of topologies, that is, points close to each other in $X$ should be close to each other in $Y$~\cite{pmlr-v162-barannikov22a}. Under this paradigm, the Representation Topology Divergence (RTD) seeks for similarities in the graphs induced by connecting points within $X$ and $Y$ that are closer than a threshold $\epsilon$. This operation is performed for all values of $\epsilon$ using a stochastic algorithm, and RTD is reported as the mean value for all measurements.
 
\subsection{K-neighborhood similarity}
Some representation similarity metrics are based on finding the $k$ nearest neighbors for each point in clouds $X$ and $Y$, and then computing the average intersection between the neighborhoods of corresponding points using Jaccard similarity~\cite{Gwilliam_2022_CVPR,Hryniowski2020} or average intersection~\cite{huh2024platonicrepresentationhypothesis}. It has been observed that larger neighborhoods have typically larger intersection due to chance~\cite{Hryniowski2020}, yet previous methods typically lack proper normalization to correct for this bias. Furthermore, these approaches require fixing $k$ to a specific scale; for example, the Platonic Representation Hypothesis~\cite{huh2024platonicrepresentationhypothesis} arbitrarily sets  $k=10$ to verify model convergence. While this allows targeting local similarity, relying on a fixed scalar obscures structural misalignments that may arise at other scales.


\subsection{Our Contribution}
Our work expands upon these foundations by:
\begin{itemize}
    \item \textbf{Normalizing} the similarity metric against a rigorous hypergeometric baseline to mitigate statistical bias;
    \item Treating the neighborhood size not as a hyperparameter, but as a continuous \textbf{spectrum}, allowing us to decouple local jitter from global structural alignment.
\end{itemize}

As a consequence, we reveal that:

\begin{itemize}
    \item The neighborhood similarity spectrum can distinguish between different distortion types—specifically decoupling \textbf{local geometric jitter} from \textbf{global structural shuffling}—which remain indistinguishable under scalar metrics;
    \item Similarity is inherently \textbf{scale-dependent}: our experiments with LLMs demonstrate that models often exhibit strong structural alignment at the semantic class level (meso-scale) while maintaining significant divergence at the instance level (micro-scale), identifying the precise granularity where convergence occurs.
\end{itemize}

\section{Topological Alignment Spectrum}
The choice of the neighborhood size $k$ is important as it changes the meaning of neighborhood-based metrics. For a low $k$, these metrics measure the local neighborhoods, which usually relates to nearest-neighborhood-based tasks such as retrieval. Higher values of $k$ can be related to more global transformations between the point clouds, but their meanings depend on the structural transformations that relate them.

It is hard to pinpoint a single value of $k$ that is more adequate in general. Because of that, we propose to calculate the measure for all relevant values of $k$. effectivelly generating a spectrum that indicates different transformations in the embedding domains. This requires an important normalization step, as discussed next.

\subsection{Normalizing the Mean Jaccard Similarity}
\label{sec:normalization}

Let $X = \{x_i\}$ and $Y = \{y_i\}$, $i \in \{1, 2, \cdots, n\}$, be point clouds where $x_i \in \mathbb{R}^{p_1}$ and $y_i \in \mathbb{R}^{p_2}$ are corresponding points. Let $N_{X, k}(x_i)$ be the set of indexes of the $k$ points closest to $x_i$ in $X$, and define $N_{Y, k}(y_i)$ analogously. Importantly, the distance measure used for this operation can be selected according to the problem. The structural similarity between corresponding points $x_i$ and $y_i$ is defined as the Jaccard similarity between their sets of indexes for nearest-neighbors:
\begin{equation}
J(N_{X,k}(x_i), N_{Y,k}(y_i)) = \frac{ |N_{X,k}(x_i) \cap N_{Y,k}(y_i)| } {|N_{X,k}(x_i) \cup N_{Y,k}(y_i)|}.
\label{eq:jaccard_similarity_AB}
\end{equation}

The mean Jaccard similarity between point clouds $X$ and $Y$ for a neighborhood size $k$ can be defined as:
\begin{equation}
    S(X,Y,k) = \frac{1}{n} \sum_{i=1}^n J(N_{X,k}(x_i), N_{Y,k}(y_i)).
    \label{eq:average_similarity_graph}
\end{equation}

The mean Jaccard similarity $S(X,Y,k)$ between point clouds depends only on their neighborhood structure: the precise definitions of distance in each domain, or the  value of point coordinates only impact the structural similarity if they change the ranks of closest points among the cloud. For instance, if the point clouds are equal, \emph{i.e} $X=Y$, then $S(X,Y,k)=1$, $\forall k \in [1,n-1]$. Furthermore, $S(X,Y,k)=S(X,Y',k)$ if $Y'$ is constructed from $Y$ by applying only isotropic scaling, translations, or orthonormal transformations, as they not change the ranks of neighborhood distances between points. The Mean Jaccard similarity $S(X, Y, k)$ has been referred to as Nearest Neighborhood Graph Similarity~\cite{Gwilliam_2022_CVPR} or as Nearest Neighborhood Topological Similarity~\cite{Hryniowski2020}. 

The neighborhood size $k$ allows explicitly adjusting the mean Jaccard similarity $S(X,Y,k)$ to evaluate similarities at different scales. However, higher values for $k$ imply an increasing chance that some elements are found in $ |N_{X,k}(x_i) \cap N_{Y,k}(y_i)|$ due to randomness, which makes $S(X,Y,k)$ not comparable along different values for $k$. We propose to normalize $S(X,Y,k)$ to a value $S'(X,Y,k)$ in which, for any $k$, a value of zero means that clouds are uncorrelated at that scale, while $1$ means that the clouds are strictly equal topologically.

For such, let $X = \{x_i\}$ and $Y = \{y_i\}$ be independent point clouds with arbitrary distributions. In this case, the neighborhoods $N_{X,k}(x_i)$ and $N_{Y,k}(y_i)$ become random draws. The intersection $N_{X,k}(x_i) \cap N_{Y,k}(y_i)$ can be seen as marking $k$ objects of interest as the ones belonging to $N_{X,k}(x_i)$, and then obtaining the intersection as $k$ random draws without replacement among the $n-1$ elements in ${y_j, j \neq i}$, with $k$ elements of interest. Hence, the intersection cardinality $|N_{X,k}(x_i) \cap N_{Y,k}(y_i)|$ for a randomly chosen $i$ follows a hypergeometric distribution with $n-1$ total elements, $k$ elements of interest and $k$ draws, that is:
\begin{equation}
\begin{array}{rl}
\mathbb{E}[|N_{X,k}(x_i) \cap N_{Y,k}(y_i)] &= \frac{\text{\# objects of interest} \times \text{\# draws}}{\text{\# total objects}}\\ &= \frac{k^2}{n-1}.
\end{array}
\label{eq:hypergeometric}
\end{equation}

We can transform Equation \ref{eq:jaccard_similarity_AB} to:
\begin{equation}
J(N_{X,k}(x_i), N_{Y,k}(y_i)) = \frac{|N_{X,k}(x_i) \cap N_{Y,k}(y_i)|}{2k - |N_{X,k}(x_i) \cap N_{Y,k}(y_i)}.
\label{eq:jaccard_with_k}
\end{equation}

Let $x = |N_{X,k}(x_i) \cap N_{Y,k}(y_i)|$ and $f(x)=J(N_{X,k}(x_i), N_{Y,k}(y_i), k)$.

This function is convex. Thus, by Jensen's inequality,

\begin{equation}
\mathbb{E}[f(x)] \geq \frac{\mathbb{E}[x]}{2k - \mathbb{E}[x]}
\end{equation}

Using the expected value of the hypergeometric distribution, we have:
\begin{equation}
\mathbb{E}[J(N_{X,k}(x_i), N_{Y,k}(y_i))] \geq \frac{k^2/(n-1)}{2k-(k^2/(n-1))}
\label{eq:expectedJ}
\end{equation}

Therefore, we can find a probabilistic lower bound for the expected value of $J(N_{X,k}(x_i),N_{Y,k}(y_i))$ if the point clouds are i.i.d:
\begin{equation}
\mathbb{E}[J(N_{X,k}(x_i), N_{Y,k}(y_i)] \geq H(k) = 
\frac{k}{2(n-1)-k}.
\label{eq:hypergeometric_lowerbound}
\end{equation}

Finally, we can use this normalization factor to define the Topological Alignment as:
\begin{equation}
\text{TA}(X, Y, k) = \frac{S(X, Y, k)-H(k)}{1-H(k)}.
\label{eq:nngs}
\end{equation}

$\text{TA}(X,Y,k)$ is consistently equal to $1$ if $X$ and $Y$ are equivalent, and equal to $0$ if $X$ and $Y$ are uncorrelated. Importantly, $H(k)$ is not a strict lower bound for $S(A,B,k)$, as there are point cloud configurations, like manifold topologies, in which points that are further away within the manifold can be closer together in the metric space. As a consequence, $\text{TA}(X, Y, k)$ can reach values below zero. Negative values indicate neighborhood overlap lower than the random baseline at that scale and should be interpreted as active anti-alignment rather than noise.

By sweeping TA for all values of $k \in [1, n-1]$, we find the Topological Alignment Spectrum (TAS). Due to the normalization, the values of TAS are comparable for all neighborhood sizes, hence they can be used to visualize the distortions between $X$ and $Y$ at different scales, which brings forward the geometric process underlying their differences. 

\subsection{Sampling datasets}
\label{sec:downsampling}
When the hypergeometric expected value $H(k)$ in Equation~\ref{eq:hypergeometric_lowerbound} is applied to large datasets, we have $n-1 \approx n$, yielding
\begin{equation}
H(k) = \frac{k}{2n-k}.
\label{eq:hypergeometric_lowerbound_large_n}
\end{equation}

In this regime, the neighborhood size can be expressed as a fraction $\alpha$ of the dataset size, $k = \alpha n$, which gives
\begin{equation}
H(k) = \frac{\alpha n}{2n - \alpha n} = \frac{\alpha}{2 - \alpha}.
\label{eq:hypergeometric_lowerbound_alpha}
\end{equation}

Therefore, for sufficiently large $n$, the expected value $H(k)$ depends only on the ratio $\alpha = k/n$, and not on $k$ and $n$ independently. Since the Topological Alignment $TA(X,Y,k)$ is obtained by normalizing the mean Jaccard similarity using $H(k)$, this normalization is invariant under proportional scaling of $k$ and $n$.

\begin{figure}[h!]
\includegraphics[width=\columnwidth]{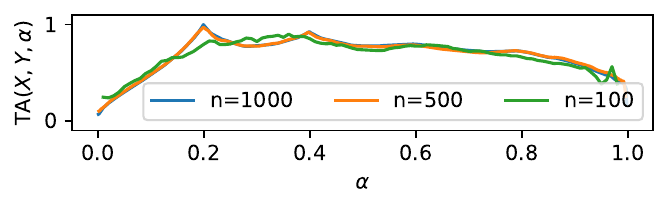}
\caption{TAS curves computed between a synthetic Gaussian point cloud and a perturbed version with added isotropic noise are shown for the full dataset and for uniformly downsampled subsets, with neighborhood sizes scaled proportionally. When plotted against the normalized neighborhood fraction 
$alpha = k/n$, the resulting spectra closely align. This illustrates approximate invariance of the TA normalization under downsampling.}
\label{fig:downsampling}
\end{figure}

As a consequence, if a dataset is uniformly downsampled and the neighborhood size is scaled proportionally (i.e., $k' = \alpha n'$), the resulting $TA$ values are preserved in expectation, provided that the local neighborhood structure is approximately maintained. This allows TAS to be computed on appropriately downsampled datasets with minimal distortion, substantially reducing computational cost. In practice, we recommend stratified sampling when strong class imbalance or small semantic clusters are present. This property, demonstrated in Figure~\ref{fig:downsampling}, is similarly explored in T-SNE~\cite{tsne2008} and UMAP~\cite{mcinnes2020umapuniformmanifoldapproximation}.

Next, we demonstrate how to apply this principle, and compare it with using scalar metrics.

\section{Using TAS to find structural similarities}
\label{sec:demo}

In this section, we show how to use TAS to identify the specific geometric processes that relate two point clouds. We contrast this with scalar metrics, showing that they cannot be used to identify these differences. For such, we artificially generated the following point cloud pairs, all with $n=500$ points:

\begin{itemize}
\item \textbf{Local Jitter}: A dataset $X$ with four equal-sized, well-separated clusters. $Y$ is generated by adding a small amount of random noise to $X$, thus reducing local neighborhood intersections, but not changing the cluster locations;
\item \textbf{Shuffled Centers}: A dataset $X$ with four equal-sized, well-separated clusters. $Y$ is generated by shuffling the cluster centers, but preserving their local neighborhoods;
\item \textbf{Random Noise}: A dataset $X$ is generated using samples from a Normal distribution. $Y$ is generated drawing the same number of samples from that same distribution.
\end{itemize}

\begin{figure}[h!]
    \centering
    \includegraphics[width=\columnwidth]{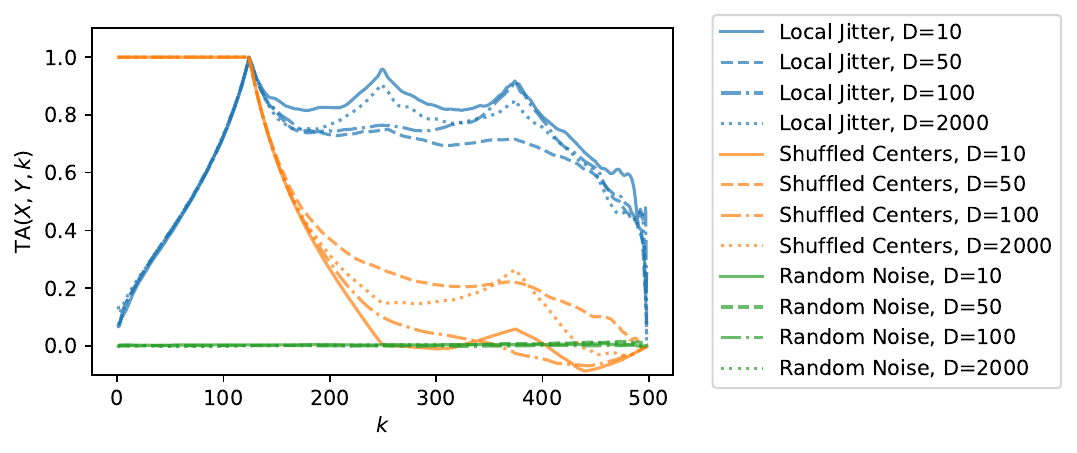}
    \caption{TAS for different distortions. Adding local jitter to a cluster topology harms the local neighborhoods, but the similarity for $k$ equal to the cluster size remains high. Shuffling cluster centers does not harm local similarity. Random noise is consistently found to have no similarity.}
    \label{fig:toy_distortions}
\end{figure}

The experiments were conducted using dimensions $D \in \{10, 50, 100, 2000\}$. The TA spectra are shown in Figure~\ref{fig:toy_distortions}. The similarity values for the scalar metrics are shown in Table~\ref{tab:toy_distortions}. The divergences (GW and RTP) were scaled to $1/(1+x)$ so that they can be interpreted as similarities between $0$ and $1$. 

\begin{table}[h!]
\centering
\begin{tabular}{c lrrrr}
\toprule
 & & \multicolumn{4}{c}{Point cloud dimension} \\
 & & 10 & 50 & 100 & 2000 \\
\midrule
\multirow{13}{*}{\rotatebox{90}{Local Jitter}}
& CKA Linear & 1.00 & 1.00 & 1.00 & 1.00 \\
& CKA RBF ($\alpha=0.2$) & 0.99 & 1.00 & 1.00 & 1.00 \\
& CKA RBF ($\alpha=0.4$) & 1.00 & 1.00 & 1.00 & 1.00 \\
& CKA RBF ($\alpha=0.8$) & 1.00 & 1.00 & 1.00 & 1.00 \\
& GULP & 1.00 & 1.00 & 1.00 & 1.00 \\
& Procrustes & 0.99 & 0.99 & 0.99 & 1.00 \\
& GW & 0.98 & 0.87 & 0.91 & 0.36 \\
& PWCCA & 0.90 & 0.89 & 0.90 & - \\
& RTD & 0.14 & 0.11 & 0.11 & 0.09 \\
\cline{2-6}
& TA ($k=10$) & 0.14 & 0.16 & 0.16 & 0.18 \\
& TA ($k=125$) & 0.99 & 0.99 & 0.99 & 0.99 \\
& TA ($k=250$) & 0.96 & 0.75 & 0.76 & 0.90 \\
\midrule
\multirow{13}{*}{\rotatebox{90}{Shuffled Centers}}
& CKA Linear & 0.71 & 0.99 & 0.97 & 1.00 \\
& CKA RBF ($\alpha=0.2$) & 1.00 & 1.00 & 1.00 & 1.00 \\
& CKA RBF ($\alpha=0.4$) & 0.97 & 1.00 & 1.00 & 1.00 \\
& CKA RBF ($\alpha=0.8$) & 0.76 & 0.99 & 0.98 & 1.00 \\
& GULP & 0.96 & 1.00 & 1.00 & 1.00 \\
& Procrustes & 0.78 & 0.99 & 0.99 & 1.00 \\
& GW & 0.99 & 0.97 & 0.99 & 0.98 \\
& PWCCA & 1.00 & 1.00 & 1.00 & - \\
& RTD &0.71 & 0.91 & 0.91 & 0.97 \\
\cline{2-6}
& TA ($k=10$) & 1.00 & 1.00 & 1.00 & 1.00 \\
& TA ($k=125$) & 0.98 & 0.98 & 0.98 & 0.98 \\
& TA ($k=250$) & 0.00 & 0.26 & 0.10 & 0.15 \\
\midrule
\multirow{13}{*}{\rotatebox{90}{Random Noise}}
& CKA Linear & 0.02 & 0.09 & 0.17 & 0.80 \\
& CKA RBF ($\alpha=0.2$) & 0.96 & 1.00 & 1.00 & 1.00 \\
& CKA RBF ($\alpha=0.4$) & 0.21 & 0.70 & 0.85 & 0.99 \\
& CKA RBF ($\alpha=0.8$) & 0.04 & 0.14 & 0.24 & 0.86 \\
& GULP & 0.84 & 0.96 & 0.98 & 1.00 \\
& Procrustes & 0.01 & 0.07 & 0.14 & 0.88 \\
& GW & 0.64 & 0.50 & 0.50 & 0.50 \\
& PWCCA & 0.11 & 0.27 & 0.40 & - \\
& RTD & 0.02 & 0.03 & 0.04 & 0.13 \\
\cline{2-6}
& TA ($k=10$) & 0.00 & -0.00 & 0.00 & 0.00 \\
& TA ($k=125$) & 0.00 & 0.00 & 0.00 & 0.00 \\
& TA ($k=250$) & 0.00 & -0.00 & 0.00 & 0.00 \\
\bottomrule
\end{tabular}
\caption{Similarities for various types of distortions, point dimensionality, and methods. }
\label{tab:toy_distortions}
\end{table}

\textbf{Local Jitter} is consistently identified by a low TA for closer neighborhoods (small $k$), with a growth until $k$ reaches the cluster size. This shows that neighborhoods were changed within each cluster, but the cluster structures were preserved, as expected. Higher values of $k$ capture a similarity that depends on the exact positioning of the clusters in each case.

The results in Table~\ref{tab:toy_distortions} show that all scalar position- and distance-based metrics consistently consider clouds distorted by local jitter as highly similar to their undistorted counterparts, with the exception of GW at dimension 2000. This is because the changes induced by local jitter in position and point-wise distance are small when compared to the inter-cluster distances. Hence, position- or distance-based metrics are prone to ignore this type of distortion.

However, we observe that RTD yields values close to a low-k NNGS.

\textbf{Shuffled Centers} are related to a complete similarity ($\text{TA}=1.0$) for neighborhoods smaller than the cluster size. Higher values of $k$ cause TA to account for which clusters are preserved, leading to a decrease in similarity. Importantly, TA reaches values below $0$ because $S(X,Y,k)$ has reached particular configurations in which the average neighborhood intersection becomes lower than $k^2/n$.

In this type of distortion, CKA, Procrustes, and RTD yield inconsistent results which grow along the point cloud dimension. GULP correctly identifies that the respositioning of point clouds does not change the predictive power of the embeddings, and PWCCA shows that shuffling centers still allow both clouds to be mapped to the same space without relevant distortion. GW shows that most distances that are changed in the shuffling algorithm can be recovered using a transport matrix.

\textbf{Random noise} is consistently found to have zero TA. This validates the use of $H(k)$ as an expected value for $S$ with random point clouds. GW also gives consistent results.

However, we note that CKA, Procrustes, PWCCA and GULP yield increasing values for larger point cloud dimensions. This indicates that they can be misleading when used to analyze high-dimensional points, and that their absolute numbers must account for the dimension changes.

We note that position-based and distance-based scalar metrics indicate a high similarity for the Local Jitter and the Shuffled Centers experiments, while RTD (topology-based) indicates the opposite. The TA spectra were able to reveal that the similarities and differences in each experiment were due to different geometric processes.

Also, we highlight that TA exhibits robustness to changes in point-cloud dimensionality, that is, its behavior remains constant regardless of the point cloud dimension. Results in Table~\ref{tab:toy_distortions} show that CKA, GULP, Procrustes, and RTD are highly susceptible to yield higher values for point clouds in higher dimensionalities. Importantly, the high values for GULP and GW indicate, respectively, that both point clouds have similar approximation capability, and that it is easy to map one cloud to another.

\section{Case study: MultiBERTs}
\label{sec:multiberts}

The results shown in Section~\ref{sec:demo} demonstrate how to interpret TA spectra in different situations. The demonstrations shows that TAS is able to identify similarities at different scales. We further conduct experiments to show how this reflects in practical applications.

For such, we used two pre-trained BERT~\cite{devlin-etal-2019-bert} models from the MultiBERTs~\cite{multiberts} collection. Models have the same architecture and were trained in the same task, but with different parameter initialization values. We fine-tuned each of the pre-trained models to the train subset of three different standard datasets (SST2, IMDB, and AG-News).

\begin{figure*}[ht]
  \centering
  \begin{subfigure}[t]{0.325\textwidth}
    \centering
    \includegraphics[width=\textwidth]{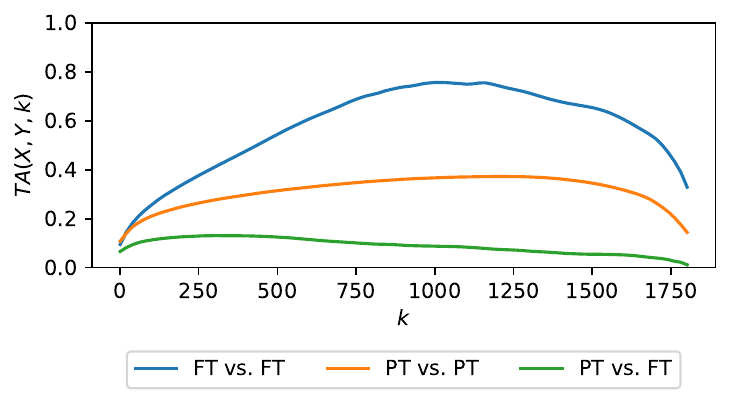}
    \caption{SST2}
    \label{fig:sst2_ptft}
  \end{subfigure}
  \hfill
  \begin{subfigure}[t]{0.325\textwidth}
    \centering
    \includegraphics[width=\textwidth]{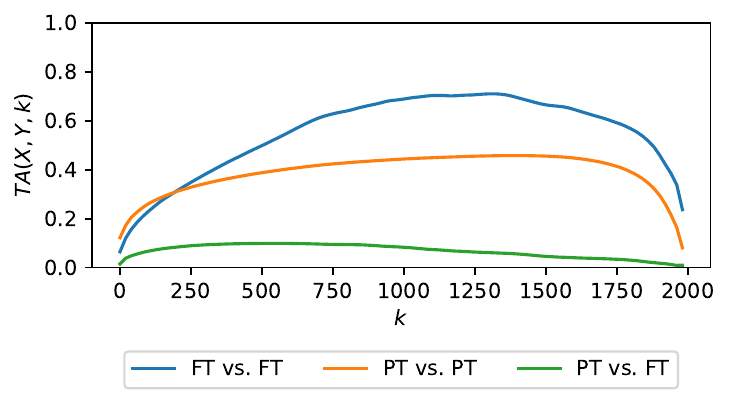}
    \caption{IMDB}
    \label{fig:imdb_ptft}
  \end{subfigure}
  \hfill
  \begin{subfigure}[t]{0.325\textwidth}
    \centering
    \includegraphics[width=\textwidth]{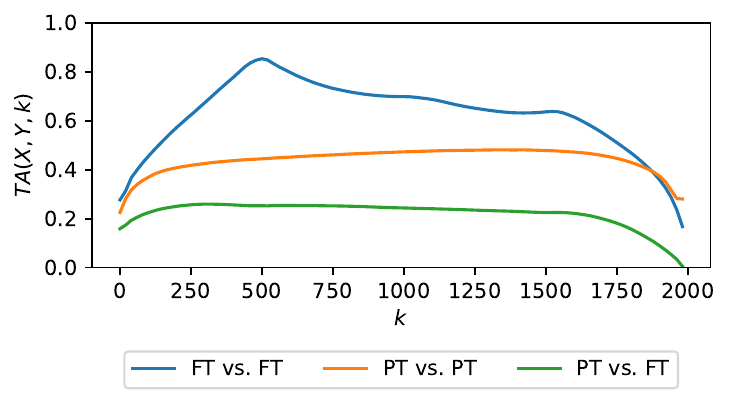}
    \caption{AG-News}
    \label{fig:agnews_ptft}
  \end{subfigure}

  \caption{TA spectra (with cosine distances) using BERT models pre-trained using different random initialization seeds and fine-tuned for each dataset. The similarity between the fine-tuned models consistently starts low and peaks at around the cluster size, indicating a behavior similar to clusters with local jitter.}
  \label{fig:ptft}
\end{figure*}

We used TAS and scalar metrics to measure the embedding similarities in their corresponding test subsets and in three situations: between the fine-tuned models (FT vs. FT), between a pre-trained model and its fine-tuned version (PT vs. FT), and between the off-the-shelf pre-trained models (PT vs. PT). The TA spectra are show in Figure~\ref{fig:ptft}, and the scalar metrics are shown in Table~\ref{tab:ptft}.

\begin{table}[h!]
\centering
\begin{tabular}{clrrr}
\toprule
 & & FT vs. FT & PT vs. PT & PT vs. FT \\
\midrule
\multirow{9}{*}{\rotatebox{90}{SST2}}
& CKA Linear & 0.93 & 0.64 & 0.23 \\
& CKA RBF & 0.92 & 0.63 & 0.22 \\
& GULP & 0.98 & 0.99 & 0.88 \\
& Procrustes & 0.73 & 0.49 & 0.31 \\
& GW & 0.99 & 0.99 & 0.78 \\
& PWCCA & 0.85 & 0.81 & 0.84 \\
& RTD & 0.06 & 0.06 & 0.01 \\
& TA ($k=10$) & 0.12 & 0.12 & 0.07 \\
& TA ($k=910$) & 0.74 & 0.36 & 0.09 \\
\midrule
\multirow{9}{*}{\rotatebox{90}{IMDB}}
& CKA Linear & 0.92 & 0.76 & 0.20 \\
& CKA RBF & 0.91 & 0.68 & 0.19 \\
& GULP & 0.98 & 0.99 & 0.88 \\
& Procrustes & 0.69 & 0.54 & 0.23 \\
& GW & 0.99 & 0.99 & 0.70 \\
& PWCCA & 0.80 & 0.83 & 0.80 \\
& RTD & 0.07 & 0.05 & 0.01 \\
& TA ($k=10$) & 0.09 & 0.14 & 0.03 \\
& TA ($k=1000$) & 0.69 & 0.44 & 0.08 \\
\midrule
\multirow{9}{*}{\rotatebox{90}{AG-News}}
& CKA Linear & 0.96 & 0.82 & 0.59 \\
& CKA RBF & 0.97 & 0.77 & 0.46 \\
& GULP & 0.99 & 0.99 & 0.92 \\
& Procrustes & 0.82 & 0.58 & 0.39 \\
& GW & 0.99 & 1.00 & 0.66 \\
& PWCCA & 0.83 & 0.81 & 0.82 \\
& RTD & 0.08 & 0.06 & 0.01 \\
& TA ($k=10$) & 0.27 & 0.25 & 0.16 \\
& TA ($k=500$) & 0.85 & 0.44 & 0.25 \\
\bottomrule
\end{tabular}
\caption{Scalar similarity measures in the finetuning experiment.}
\label{tab:ptft}
\end{table}

\subsection{Fine-tuned models generate embedding spaces similar to clusters with local jitter}

The comparison between two fine-tuned models in Figure~\ref{fig:ptft} shows patterns that resemble the Local Jitter experiment in Figure~\ref{fig:toy_distortions}: a low similarity for low $k$, which increases until $k$ reaches the cluster size, followed by another decrease. This indicates that classifiers create clusters with similar content, but different structure. As indicated by GULP, these clusters have similar prediction power, and, as indicated by GW, they are consistently easy to map from one another.

TA spectra was the only measure able to capture the differences between the local and the cluster-level structures. Using an RBF kernel in CKA was not effective for such. This indicates that neighborhoods have a more predictable statistical behavior (as captured by TA) than distances (as captured by CKA), which makes them more adequate to evaluate embedding similarities at different scales.

\subsection{Pre-trained models with different seeds span different geometries}

The comparison between different pre-trained models show that they yield significantly different geometries, which is evidenced by low values of TA, CKA, and Procrustes. The TA spectra reveal that the differences are higher in the immediate neighborhood. The high values for GULP and GW indicate that, although different, these models can have similar predictive power and can be easily mapped to one another.

\subsection{Fine-tuning changes neighborhoods in all scales}

Figure~\ref{fig:ptft} show a consistent pattern in which the TA between the pre-trained model and its finetuned counterpart are low throughout the spectra. This indicates that fine-tuning greatly changes the yielded representation geometries, which is corroborated by CKA and Procrustes measures shown in Table~\ref{tab:ptft}. Moreover, the PT vs. FT case is the one in which GULP and GW yield lower values, which shows this large change impacts both the predictive power and the distances among points.

\section{Discussion}

The flexibility of changing the neighborhood size in TA, leading to the construction of the TA spectrum, can diagnose the nature of the geometric transformations that happen from one cloud to the other. It allows differentiating local transformations from more global ones. To the best of our knowledge, this is the first work to normalize mean Jaccard overlap against its hypergeometric lower-bound expected value and sweep it as a diagnostic spectrum.

The results in Section~\ref{sec:multiberts} further add to previous work on the geometry of fine-tuning. It has been previously found that fine-tuning increases the distance between clusters~\cite{zhou-srikumar-2022-closer}, which is corroborated by the high TA at a $k$ equal to the cluster size. Moreover, it has been found that layers seldom change during fine-tuning~\cite{zhou-srikumar-2022-closer} and preserve their language representation power~\cite{merchant-etal-2020-happens}, but the PT-FT curves in Figure~\ref{fig:ptft} indicate that these seemingly small changes can nevertheless induce substantial topological reorganization in the resulting representations.

\subsection{Computational considerations}

The computation of TAS relies on nearest-neighbor queries and set intersections across multiple neighborhood sizes. In practice, all neighborhoods up to a maximum size $k_{\max}$ can be obtained from a single $k_{\max}$-nearest-neighbor computation, and Jaccard intersections can be computed efficiently in batch using vectorized tensor operations. Furthermore, as shown in Section~\ref{sec:downsampling}, TAS is approximately invariant under proportional downsampling, allowing large datasets to be analyzed using representative subsets with scaled neighborhood sizes.

\subsection{Limitations}

TAS characterizes representational similarity through neighborhood preservation and is therefore most informative when neighborhood structure is meaningful and sufficiently stable. In extremely small datasets, highly sparse regimes, or representations dominated by noise, neighborhood statistics may become unstable and TAS values should be interpreted with caution. Moreover, TAS captures structural alignment rather than functional equivalence: representations with identical predictive behavior but different internal geometries may exhibit low topological alignment.




\section{Conclusion}

Traditional representation similarity indices provide a limited ``temperature reading'' of neural alignment, often failing to detect the precise granularity where models diverge. The introduction of the Topological Alignment Spectrum (TAS) shifts representational analysis from point estimates to continuous spectra, offering a robust defense against the biases of high-dimensional latent spaces.

The empirical evidence provided here marks a departure from the ``localized'' adjustment" theory of fine-tuning. TAS demonstrates that the topological pressure of task-specific data overrides pre-trained ancestry, triggering a comprehensive structural reorganization that scalar metrics like CKA and GULP are fundamentally limited in their ability to resolve. As neural architectures continue to scale, the transition from global scalars to topological spectra will be essential for ensuring the structural reliability of deep learning systems.

\section{Acknowledgments*}

We used Gemini to polish the writing in this text.

\bibliographystyle{IEEEtran}
\bibliography{refs}

\end{document}